\documentclass{article} 
\usepackage{colm2024_conference}

\usepackage{microtype}
\usepackage{hyperref}
\usepackage{url}
\usepackage{booktabs}
\usepackage{graphicx}
\usepackage{multirow}
\usepackage{multicol}

\title{An Image is Worth 1/2 Tokens After Layer 2: Plug-and-PLay Acceleration for VLLM Inference}

\colmfinalcopy

\author{Liang Chen$^1$, Haozhe Zhao$^1$, Tianyu Liu$^{2\dagger}$, Shuang Bai$^2$, Junyang Lin$^2$\\\textbf{Chang Zhou$^2$, Baobao Chang$^{1\dagger}$}\\ $^1$National Key Laboratory for Multimedia Information Processing, Peking University\\$^2$Alibaba Group\\
\texttt{leo.liang.chen@outlook.com}
}

\begin{document}

\maketitle

\def\thefootnote{$^\dagger$}\footnotetext{Corresponding author.}
\def\thefootnote{\arabic{footnote}}

\begin{abstract}
In this study, we identify the inefficient attention phenomena in Large Vision-Language Models (LVLMs), notably within prominent models like LLaVA-1.5, QwenVL-Chat, and Video-LLaVA. We find that the attention computation over visual tokens is extremely inefficient in the deep layers of popular LVLMs, suggesting a need for a sparser approach compared to textual data handling. To this end, we introduce FastV, a versatile plug-and-play method designed to optimize computational efficiency by learning adaptive attention patterns in early layers and pruning visual tokens in subsequent ones.
Our evaluations demonstrate FastV's ability to dramatically reduce computational costs (e.g., a 45\% reduction in FLOPs for LLaVA-1.5-13B) without sacrificing performance in a wide range of image and video understanding tasks. The computational efficiency and performance trade-off of FastV are highly customizable and Pareto-efficient. It can compress the FLOPs of a 13B-parameter model to achieve a lower cost than that of a 7B-parameter model while still maintaining superior performance. We believe FastV has practical value for the deployment of LVLMs in edge devices and commercial models.  Code is released at \href{https://github.com/pkunlp-icler/FastV}{github.com/pkunlp-icler/FastV}.


\begin{figure}[h]
\centering
\includegraphics[width=0.9\textwidth]{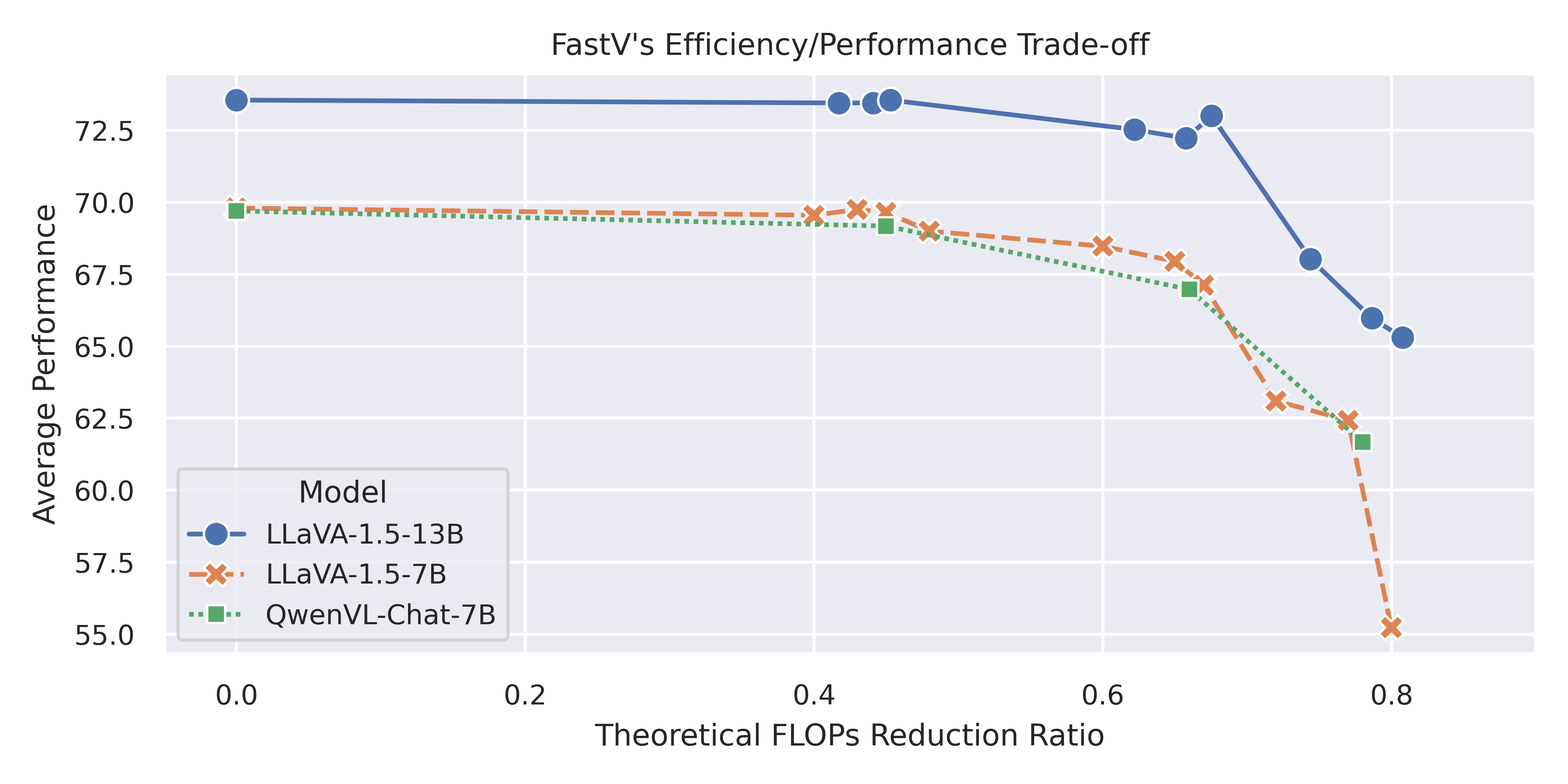}
\caption{The Efficiency/Performance trade-off curve of FastV. The x-axis stands for the theoretical FLOPs reduction ratio under different FastV configurations. The y-axis stands for performance under different settings, we report the average scores of \{Nocaps (Cider), Flickr30k (Cider), A-OKVQA (Acc), MMMU (Acc)\}. We can see that FastV can achieve 45\% FLOPs reduction with nearly no performance loss for different models.}

\label{fig:pareto_front}
\end{figure}
\end{abstract}

\newpage

\section{Introduction}

Large Vision-Language Models (LVLMs) have become a hit in both computer vision and natural language processing studies. We have witnessed tremendous creative research and applications that are built upon powerful LVLMs~\cite{liu2023llava,liu2024world,geminiteam2023gemini,Qwen-VL}. From describing the given picture to navigating the internet~\cite{zheng2024gpt4visionweb}, using smartphones~\cite{wang2024mobileagent} and making decisions in the real world~\cite{driess2023palme,chen2024pcabench}, large language models with vision abilities are reshaping how we interact with AI systems, which cannot be achieved solely by language or vision uni-modal models.

Currently, a majority of popular LVLMs rely on sequential visual representation, where images are transformed into hundreds or thousands of tokens when feeding them to LLM along with language prompts~\cite{gpt4v,zhu2023minigpt4,liu2023llava,zhao2023mmicl,Qwen-VL}. As LVLMs leverage the advanced emergent capabilities inherent in their language components, they concurrently face a surge in computational complexity, correlating with cost increments. This complexity stems from the principle that the proficiency of Large Language Models (LLMs) is predominantly influenced by their scale. Two critical areas remain under-explored in this context: 1) How do language models process and interpret images? and 2) While the efficient training and inference of LLMs have attracted considerable attention, these dimensions within LVLMs are yet to be thoroughly examined and understood.

In this paper, we uncover the fact that current LVLMs actually apply an inefficient way while processing image information.
Specifically, the image tokens receive strikingly lower attention scores compared to their textual counterparts within the token-based LVLMs like LLaVA. The degree of imbalance also varies between the shallow and deep layers. 
In the image captioning tasks, 
we observed that within the deep layers (after layer 2) of renowned LVLMs such as LLaVA 1.5, 
image tokens garner an average attention score that amounts to only 0.21\% of the score attributed to system prompts.
In contrast, this figure reaches 50\% in the initial two layers. 
These observations raise questions upon the optimal utilization of visual information within LVLMs. 

To address the problem, we assume a plausible explanation is that the high redundancy in visual signals leads to the aggregation of image-related, instruction-specific features onto certain ``anchor'' tokens through the self-attention mechanism in the shallow layers. Notably, these anchor tokens are not image tokens. In deep layers, attentions are focused on those anchor tokens, leading to significantly reduced attention on the image tokens themselves.

The phenomena inspires to propose FastV, a dynamic image tokens pruning method to reduce the inference cost of LVLMs. Our findings suggest an intriguing possibility: Given that image tokens contribute minimally to output generation in deeper layers due to diminished attention, why not consider removing them at these stages? FastV implements an image token pruning strategy at one specific layer of LLM. Prior to this layer, computations proceed as usual. Beyond this selected layer, image tokens are re-evaluated based on their average received attention scores. Tokens falling below a predefined attention score threshold are then selectively discarded in subsequent layers, streamlining the process by focusing on the most impactful tokens. 

Compared to other attention-based methods for accelerating inference, such as sparse attention, FastV's most notable distinction lies in its direct elimination of tokens. This approach not only bypasses the computational demand of the self-attention module but also the Feed-Forward Network (FFN) module in deeper layers. As a result, FastV achieves a great theoretical reduction in FLOPs while maintaining relatively high performance as shown in Figure~\ref{fig:pareto_front}'s experiment on LLaVA and Qwen-VL-Chat models. Our experiment on LLaVA-1.5-13B model shows that we can filter out 50\% image tokens after layer 2 without sacrificing the average performance on a combination of Vision-Language tasks including captioning tasks like Nocaps~\cite{agrawal2019nocaps}, Flickr30K~\cite{plummer2015flickr30k}, multimple choice tasks like A-OKVQA~\cite{schwenk2022aokvqa},  MMMU~\cite{yue2023mmmu}, complex embodied reasoning task like PCA-Bench~\cite{chen2024pcabench,chen2023pcaeval}, tasks requiring detailed OCR ablitily like OCR-VQA~\cite{mishra2019ocr_vqa}, more challenging video understanding tasks~\cite{jang2017tgifqa,xu2017msrvtt_qa,xu2017msvdqa} and more fine-grained evaluation like MME~\cite{fu2023mme}, MMVet~\cite{yu2023mmvetevaluatinglargemultimodal} and SeedBench~\cite{li2023seedbenchbenchmarkingmultimodalllms}.
Our latency test experiment on A-OKVQA showed that LLaVA-13B model with FastV could achieve a lower latency than LLaVA-7B model while maintaining superior performance. This result highlights the effectiveness of FastV in balancing the trade-off between speed and accuracy in LVLMs.

Researches~\cite{liu2023llava,li2023monkey} underscore the significance of enhancing image resolution for the performance of LVLMs. However, it's equally important to note that increased resolution comes with its own challenges, including a rise in the computational costs such as longer image token sequence and inference latency. We also conduct experiments on training LVLM in different image feature resolution by setting pooling layer of different strides. Specifically, with an equal number of image tokens, models equipped with FastV can process higher resolution images, leading to better performance than models limited to lower resolution features. This finding highlights the potential to enhance downstream performance by increasing image resolution without incurring additional inference costs.

In summary, the contribution of the work are three-folds:
\begin{enumerate}
    \item Identify and analyze the inefficient visual attention phenomena in prevailing LVLMs.
    \item Propose FastV, a plug-and-play method to significantly reduce inference cost for LVLMs without sacrificing performance inspired by our observation.
    \item Validate the effectiveness of FastV on a wide range of vision-language tasks across different LVLMs with thorough ablations.
\end{enumerate}

\section{Related Work}

\paragraph{Large Vision-Language Model.}
To benefit from the advancement of LLM and integrate visual information into the LLM, large Vision-Language Models utilize a Visual Prompt Generator~\cite{li2023Cheetor} to transform the visual embeddings into prompts that the language model can comprehend~\cite{li2023blip2,liu2023llava}, resulting in a significant increase in required tokens.
Handling higher resolution images inevitably necessitates a quadratic increase in the number of needed tokens. For instance, LLAVA process 336x336 images into 576 tokens~\cite{liu2023llava15} and process images with a greater resolution of 672x672 into 2304 tokens~\cite{liu2024llavanext}.
Fuyu~\cite{fuyu-8b}, in a similar vein, translates pixel-level images of 1080x1080 into 1296 tokens.
Understanding and generating multiple images or videos also inherently demands an escalated count of tokens for vision information. Both Video-Poet~\cite{kondratyuk2023videopoet} and Unified-IO2~\cite{lu2023unifiedio2} are compelled to reserve thousands of tokens within the context to facilitate the understanding and generation of multiple images or videos. 
Large multimodal models like Gemini~\cite{geminiteam2023gemini} and LWM~\cite{liu2024world} highlights the significance of long context in developing a robust understanding of the world model and extending the context length to 1M to address the issue of escalating context requirements.



\paragraph{Inference Optimization for LLM.}
Efficient inference in LLMs is challenged by their autoregressive generation where each token prediction depends on the preceding context. 
Hence, considering the quadratic complexity of computation's attention during training, as the context length increases, the generation becomes progressively slower. 
To tackle these challenges, pioneering studies
fall into two categories: methods optimizing memory consumption for attention module like FlashAttention, vLLM and RingAttention~\cite{dao2022flashattention,dao2023flashattention2,kwon2023vllm,liu2023ringAttention}, which ensure no drastic shifts in the results, and methods like StreamingLLM and FastGen~\cite{xiao2023streamingllm,ge2024fastgen} that simplify computations by pruning redundant attention computation. We are interested in the second kind of methods since they are proposed inspired by the distinct attention patterns observed in LLM's inference.
While these methods have boosted the inference efficiency of LLMs, they are designed for text-only language models, and whether their effectiveness can be transferred to LVLMs remain under-explored. 
There is previous work attempt to handle the long-context in LVLMs efficiently, like LLaMA-VID~\cite{li2023llamavid}, which utilizes cross-attention to effectively represent each video frame with two key tokens, the requirement for an additional fine-tuning stage obstructs its broad applicability for different LVLMs.



\paragraph{Token Reduction for VLMs.} There have been studies on improving efficiency for Vision-Language Models (VLMs) before the era of large vision-language models. A majority of them focus on token reduction for vision transformers (ViTs). Various methods, such as EViT \cite{liang2022patchesneedexpeditingvision}, SPViT \cite{kong2022spvitenablingfastervision}, and Pumer \cite{cao2023pumerpruningmergingtokens}, have been proposed for ViTs. More recently, PYRA \cite{xiong2024pyraparallelyieldingreactivation} has enhanced the training and inference of ViTs via a specialized token merging technique. FastV is the first to explore visual token reduction for Large Vision-Language Models (LVLMs), which uses language as an interface for various vision-language tasks. FastV utilizes the signal from LLM to guide the pruning of visual tokens, a strategy not previously explored. We are the first to demonstrate the effectiveness of token reduction in video-QA and various comprehensive LVLM benchmarks. Another significant advantage of FastV over previous methods is its simplicity; it can be applied to any LVLM without requiring model retraining.

\section{Inefficient Visual Attention in VLLMs}
\subsection{Preliminaries}
In this section, we delve into how LVLMs process visual tokens during output generation from the perspective of self-attention module. For an image-question pair $(d,t)$, the given LVLM $M$, usually in the structure of transformer~\cite{vaswani2017attention} decoder, predicts the answer $\hat{y} = M(d,t)$ in an auto-regressive manner:

\begin{equation}
p(\hat{y})=\prod_{i=1}^N p_{M}\left(\hat{y_i} \mid \hat{y}_{1\sim i-1} ;d;t \right)
\end{equation}

\begin{figure}[!t]
\centering
\includegraphics[width=0.8\textwidth]{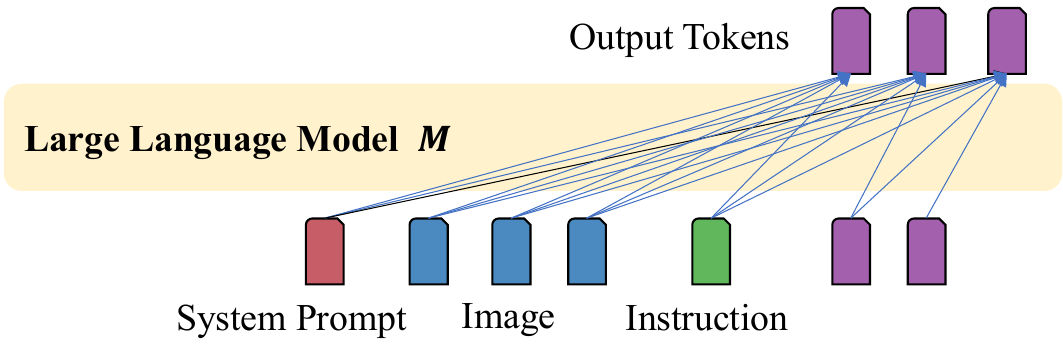}
\caption{Classic network architecture of LVLM. Image tokens and different types of text tokens are sent to the LLM as input. LLM generates output tokens conditioned on the input tokens and preceding output in an auto-regressive manner.}

\label{fig:mllm-struct}
\end{figure}

Multimodal information, encompassing both images and text, is transformed into sequential embeddings prior to being processed by the transformer model. For images, a commonly used approach is to employ a pretrained encoder, such as CLIP-VIT~\cite{radford2021clip}, to extract visual features. These features are then linearized by eliminating the spatial dimension. Additional linear transformations~\cite{zhu2023minigpt4,liu2023llava15} or cross-attention~\cite{li2023blip2,Qwen-VL} modules are utilized to adjust the size of the visual features to match the embedding size of the Large Language Model (LLM) and to achieve semantic alignment. Regarding text, a tokenizer breaks down the natural language into discrete tokens and then performs an embedding lookup to form text embeddings. In the rest of the paper, we refer to 'visual tokens' and 'text tokens' not merely as the discrete units of visual and textual data but as the embeddings derived from these units. 

As illustrated in Figure~\ref{fig:mllm-struct}, after preprocessing the image and text token to a unified embedding space, they are fed to the transformer decoder to generate output tokens. The input tokens at each decoding step can be categorized into four distinct types: system prompt (sys), image tokens (img), user instruction (ins), and output tokens (out). The system prompts for LVLMs usually inherit the backbone LLM, used as a general message to control the LLM's behavior, which is decided during the instruction tuning stage of LLM. Image tokens are the linearized image features transformed by a pretrained vision encoder. User instruction specifies the query question for the given image. Output tokens are generated step by step conditioned on the preceding tokens.

\subsection{Experiment Settings}

To explore how LVLMs process image tokens, we first randomly sample $N$ image-text pairs $D=\{(d^1,t^1),...,(d^{N},t^{N})\}$ from a combination of vision langauge tasks including image caption (Flickr30K), embodied reasoning (PCA-Bench), visual question answering (A-OKVQA), multimodal understanding and reasoning (MMMU) and then prompt the LVLM to generate $N$ responses $\hat{Y} = \{\hat{y}^1,...,\hat{y}^{N}\}$.

During the decoding process of one response, we collect each output tokens' attention score distribution $\alpha$ in different layers and sum up for different type of input tokens. That is, for the $i$-th token, in the $j$-th layer, we compute $\alpha^{i,j}_{sys}, \alpha^{i,j}_{img}, \alpha^{i,j}_{ins}, \alpha^{i,j}_{out} $ to denote the total attention score current token attends to the system prompt, image tokens, user instruction and output tokens. We have:

\begin{equation}
\alpha^{i,j}_{sys}+\alpha^{i,j}_{img}+\alpha^{i,j}_{ins}+\alpha^{i,j}_{out}=1 
\end{equation}

We compute the total attention allocation $\lambda$ to denote the total attention score one type of tokens received in one layer.  For example, the total attention of system prompt in layer $j$ is:

\begin{equation}
        \lambda_{sys}^j = \sum_{i=1}^n\alpha_{sys}^{i,j} 
\end{equation}

\begin{figure}[!t]
\centering
\includegraphics[width=\textwidth]{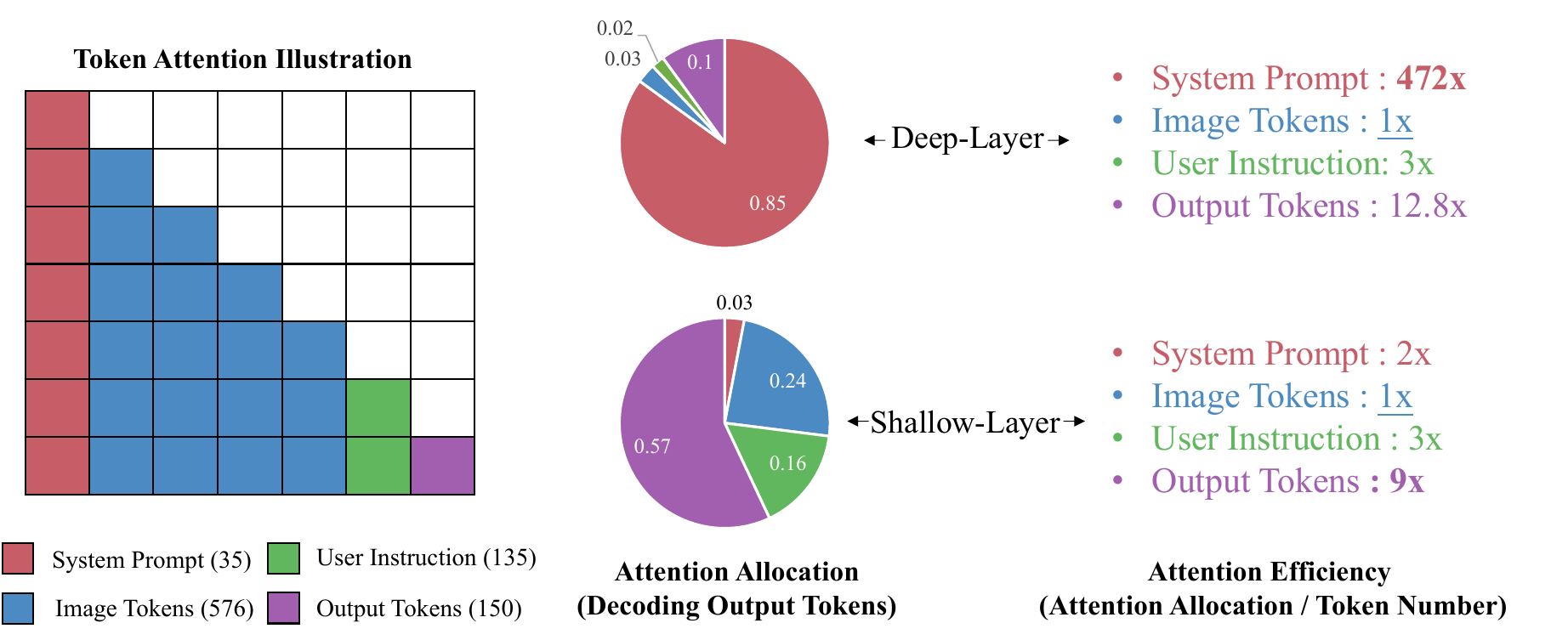}
\caption{Illustration of inefficient visual attention phenomena. The left part shows the relative position and average number of different type of input tokens, tokens could only attend to preceding tokens in the self-attention module. In average, image tokens take up most of the input tokens (64\%).   The middle and right part show the average attention allocation $\lambda$ and attention efficiency $\epsilon$ in shallow and deep layers. Image tokens receive far less attention relative to their number in the deep layers. }
\label{fig:attn-ill}
\end{figure}

where $n$ is the number of tokens in the response. Final attention allocation is averaged over all attention heads in the $N$ image-text pairs we sampled.

Next, we define metric \textbf{attention efficiency} $\epsilon$ to denote the average attention score per type's token received in one layer during the decoding process of one response. For example, the attention efficiency of image tokens in layer $j$ is:

\begin{equation}
    \epsilon_{img}^j = \frac{\sum_{i=1}^n\alpha_{img}^{i,j} }{|img|}
\end{equation}

where $|img|$ is the number of image tokens, $n$ is the number of tokens in the response. Final attention efficiency is averaged over all attention heads in the $N$ image-text pairs we sampled.

In our experiment, $N$ is set to 1000 and we use LLaVA1.5-7B as the LVLM. We follow the same generation configuration as the original paper \cite{liu2023llava}.

\subsection{Results}

We have two major findings in the attention pattern statistics regrading attention allocation $\lambda$ and attention efficiency $\epsilon$ for different type of input tokens.  We define the first 2 layers as shallow layer and the rest 30 layers as deep layers.

\begin{enumerate}
    \item Both attention allocation and attention efficiency show different degree of imbalance, which is related to the layer depth. The average attention allocation and efficiency in different layer is shown in Figure~\ref{fig:attn-ill}. In shallow layer the attention allocation is relatively more balanced than in deep layers. In shallow layer, the output tokens tends to attend to the previous output tokens while in deep layers, they tend to attend to the system prompt.

    \item Image tokens have the \textbf{lowest} attention efficiency in both shallow and deep layers. System prompt is of extremely high attention efficiency in deep layers, which is \textbf{472} times that of image tokens, taking up 85\% total attention scores.
\end{enumerate}

\begin{figure}[!t]
\centering
\includegraphics[width=\textwidth]{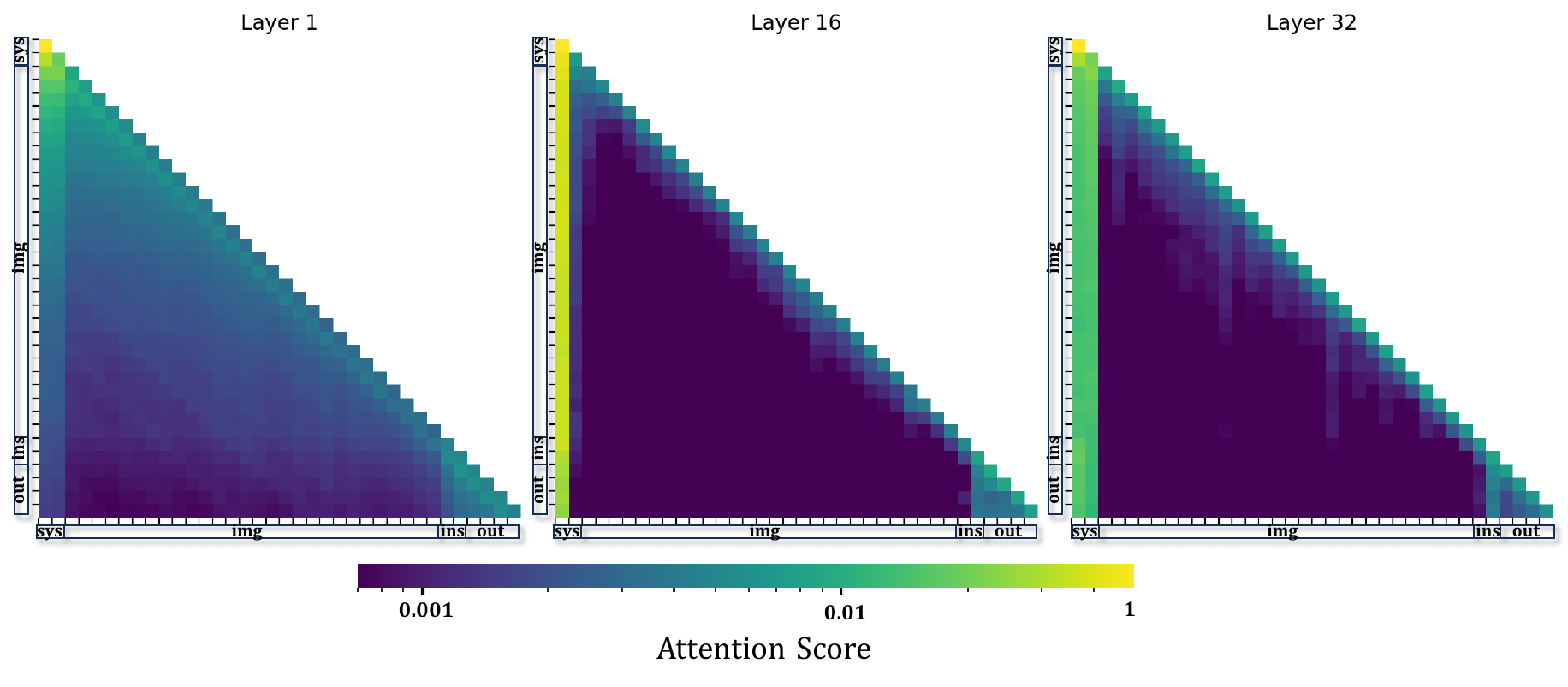}
\caption{The attention maps during the decoding process of one model response for LLaVA1.5-7B. We can see that in the bottom layer, attention distributes relatively smooth across different type of tokens. In the the deep layers, above from local attention, the attention scores are aggregated to system prompt, instruction and output tokens and attention over image tokens is rather sparse.  }

\label{fig:attn-map}
\end{figure}

\subsection{Insights}

The statistics reveal a surprising trend in the decoding process of LVLMs: despite accounting for the majority of tokens in the input, image tokens receive significantly less attention. 
Conversely, system prompts, which provides the minimal semantic information, attract the most of the attention scores. 
To delve deeper into this phenomenon, we analyze the attention maps of the first, middle, and last layers during during the decoding process of a model response as shown in Figure~\ref{fig:attn-map}. The attention maps for all layers are provided in figure-7 of the supplement material.

From the attention visualization results, we can see that in shallow layer, the attention scores distribute more smoothly across different tokens. While in deep layer, there are vertical strong lines (in the system prompt) that takes up most of attention scores. The existence of vertical strong line shows that there are some input tokens that consistently received high attention during the whole decoding process. This also explains the highly imbalanced attention efficiencies in our statistics: A small portion of anchor tokens aggregate the information from all input tokens and the model much favors to attend to those anchor tokens in deep layers. Our findings also align with the information flow of Large Language Model found in ~\cite{wang-etal-2023-label}.



\section{FastV}
With insights from the validated phenomena and explanation, we propose FastV as a solution to reduce the inference costs of LVLMs without sacrificing the performance.

\begin{figure}[!t]
\centering
\includegraphics[width=\textwidth]{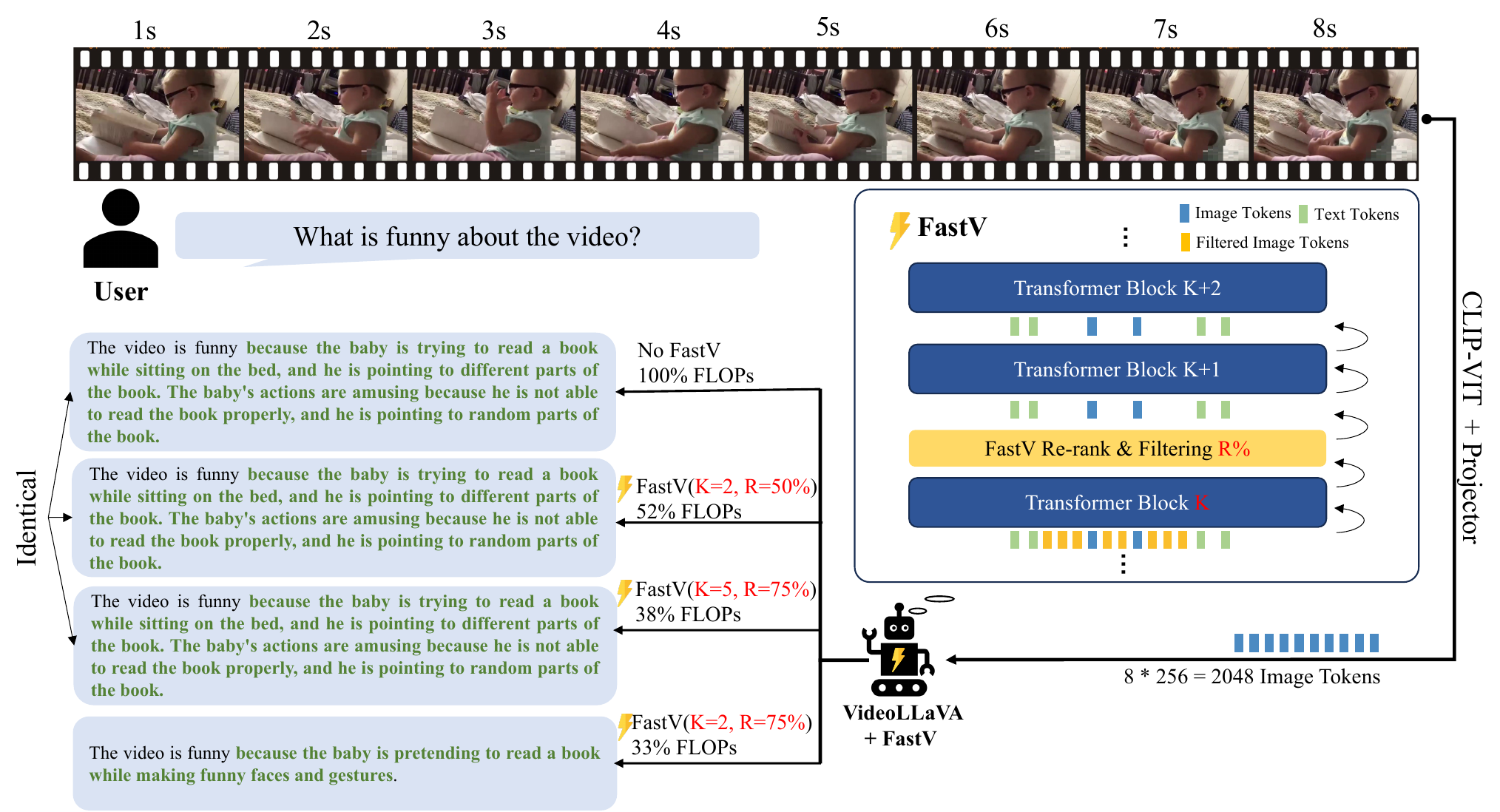}

\definecolor{mygreen}{RGB}{39, 150, 52}

\caption{Illustration of FastV. For image or video input (multiple image frames), they are first transformed to visual tokens with a pretrained image encoder like CLIP-VIT and then processed by the LLM decoder. FastV dynamically prunes $R\%$ image tokens after layer $K$ in the forward process of input tokens. We can tell from the output that FastV does not influence the correctness while reducing significant FLOPs. The correct facts in the outputs are marked \textcolor{mygreen}{\textbf{green}}. The first three outputs are completely identical.}

\label{fig:fastv}
\end{figure}

\subsection{Dynamically Prune Vision Tokens}

Figure~\ref{fig:fastv} illustrates the general idea of FastV. The key is the image token re-rank and filtering module. It consists of one ranking function $f_\phi$ and two parameters: filtering layer $K$ and filtering ratio $R\%$. At layer $K$ of the LVLM, the ranking function $f$ takes a sequence of input tokens and rank them by certain importance criteria $\phi$. The last $R\%$ tokens after ranking would be pruned out in successive layers. We simply compute the average attention-score one token received from all other tokens as the criteria $\phi_{attn}$ in our experiment. 
In extreme condition, $K$ could be also set to 0, that image tokens are pruned before sending to the language model, we use random ranking as the criteria $\phi_{rand}$ where image tokens are randomly dropped.

FastV is plug-and-play to different token-based LVLMs for various vision language tasks without the need of training the model. We take video understanding tasks with VideoLLaVA~\cite{lin2023video} as example as shown in Figure~\ref{fig:fastv}. 

\subsection{Computing Cost Estimation}


We consider the computation of multi-head attention (MHA) and feed-forward network (FFN) module in the FLOPs estimation. For one transformer layer, assume $n$ is the token number, $d$ is the hidden state size, $m$ is the intermediate size of FFN, the total FLOPs can be estimated by $4nd^2+2n^2d+2ndm$. For the whole model, assume FastV prunes tokens from $n$ to $\hat{n}=(1-R\%)\cdot n$ after layer $K$ and there are T layers at all. The theoretical FLOPs reduction ratio related to image tokens is computed as:

\begin{equation}
    1-\frac{K\times(4nd^2+2n^2d+2ndm)+(T-K)\times(4\hat{n}d^2+2\hat{n}^2d+2\hat{n}dm)}{T\times(4nd^2+2n^2d+2ndm)}
\end{equation}

\begin{figure}[h]
\centering
\includegraphics[width=0.6\textwidth,keepaspectratio]{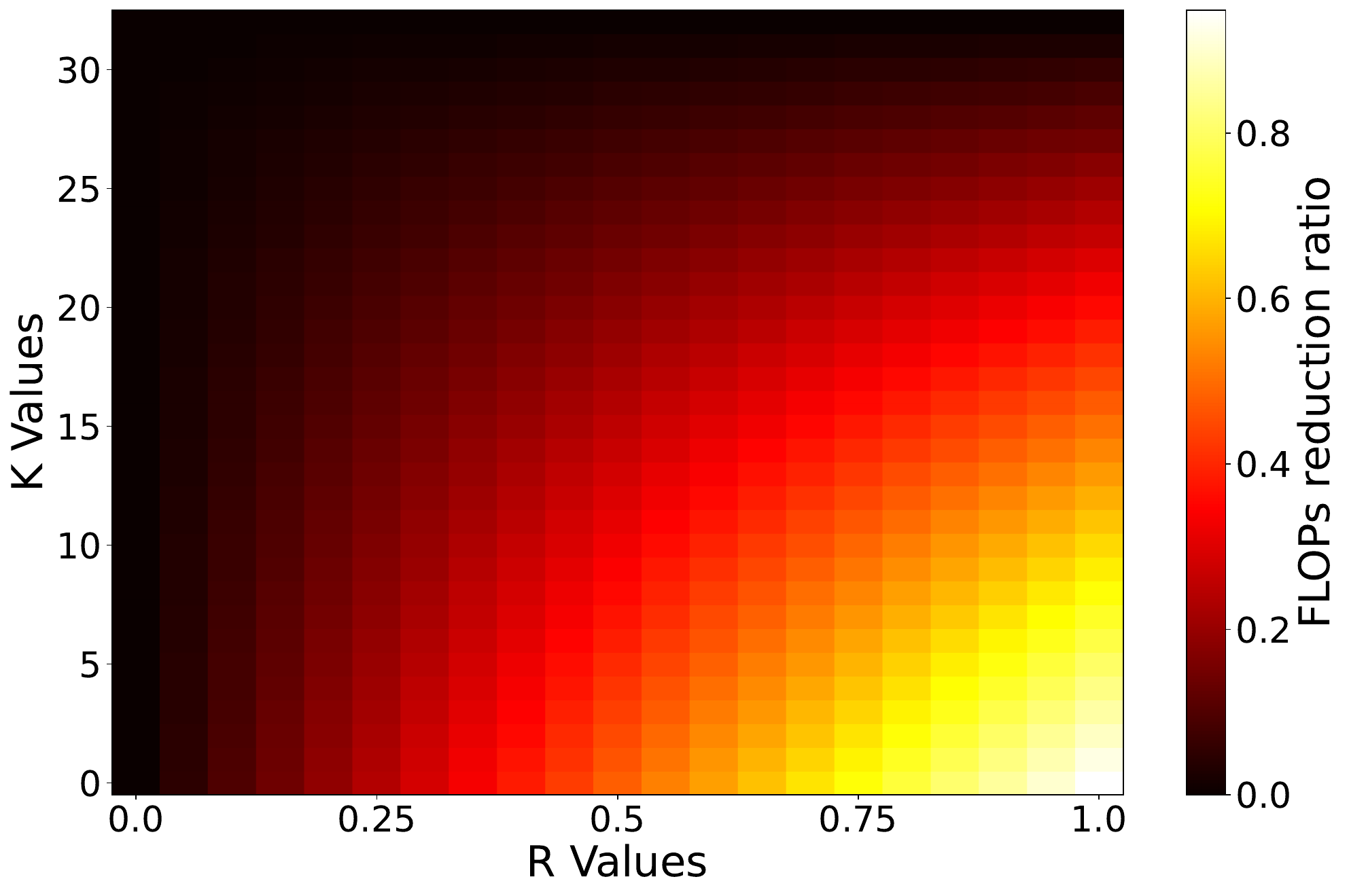}
\caption{The heat map of theoretical FLOPs reduction ratio. The color in the figure represents the reduction ratio in different $K$ and $R$ in FastV. }
\label{fig:reduction_ratio_map}
\end{figure}

We plot a 3D graph to show how the FLOPs reduction ratio changes with FastV's parameter $K$ and $R$ in Figure~\ref{fig:reduction_ratio_map}.

\subsection{Comparison: Training With Less Visual Tokens}

FastV achieves computation reduction through eliminating redundant visual tokens during inference stage. An alternative method to reduce visual tokens is directly training with less visual tokens. This could be simply done by conducting pooling on the output of visual encoder during LVLM's training process. We compare FastV and this method in our ablation studies (sec.~\ref{sec:ablation}).

\section{Experiment}
\begin{table*}[!t]
\caption{Performance/Computation Balance of FastV under different configurations (K for filtering layer, R for filtering ratio). \textcolor{red}{Highest score} for each model is in red while the \textcolor{blue}{second highest} is in blue.}
\centering
\resizebox{0.8\columnwidth}{!}{
\begin{tabular}{c|cccc|ccccc}
\toprule
\multirow{2}{*}{Model} & \multicolumn{4}{c|}{FastV Settings}  & Nocaps & Flickr30k & A-OKVQA & MMMU   & \multirow{2}{*}{Avg}\\
 ~& K & R & Flops(B) & Flops Ratio & \textit{CIDEr} & \textit{CIDEr} & \textit{Accuracy} & \textit{Accuracy}  & ~ \\ 

\midrule
\multirow{13}{*}{LLaVA-1.5-7B}
        ~ & \multicolumn{2}{c}{Baseline} & 99.3 & 100\%  & \color{blue}{99.8} & \color{blue}{67.9} &76.7 &\color{blue}{34.8} &\color{red}{\textbf{69.8}} \\
        ~ &2 & 90\% & 19.9  & 20\% & 72.1 & 43.7 & 70.1 & \color{red}{\textbf{35}} & 55.2 \\ 
        ~ &2 & 75\% & 32.8  & 33\% & 94.6 & 63.6 & 75.5 & \color{blue}{34.8} & 67.1 \\ 
        ~ &2 & 50\% & 54.6  & 55\% & 99.7 & 67.5 & \color{red}{\textbf{77}} & 34.4 & \color{blue}{69.7} \\ 
        ~ &3 & 90\% & 22.8  & 23\% & 87.2 & 55.8 & 71.9 & \color{blue}{34.8} & 62.4 \\ 
        ~ &3 & 75\% & 34.8  & 35\% & 98 & 65 & 74.7 & 34.1 & 68.0\\ 
        ~ &3 & 50\% & 56.6  & 57\% & 99.7 & \color{red}{\textbf{68.3}} & 76.7 & 34.3 & \color{red}{\textbf{69.8}} \\ 
        ~ &5 & 90\% & 27.8  & 28\% & 88.6 & 59.3 & 70.6 & 33.9 & 63.1 \\ 
        ~ &5 & 75\% & 39.7  & 40\% & 98.5 & 66.3 & 74.8 & 34.3 & 68.5 \\ 
        ~ &5 & 50\% & 59.6  & 60\% & 99.2 & \color{blue}{67.9} & \color{blue}{76.8} & 34.3 & 69.6 \\ 
        ~ &0 &90\% & 18.9 & 19\% & 7 & 53.2 & 66.8 & 34.7  & 40.4  \\
        ~ &0 &75\% & 28.8 & 29\% & 27.2 & 61.4 & 72.8 & 35.1  & 49.1  \\ 
        ~ &0 &50\% & 51.6 & 52\% & \color{red}{\textbf{100.9}} & 65.5 & 75.3 & 34.3  & 69.0 \\

\midrule
\multirow{10}{*}{LLaVA-1.5-13B} 
        ~ & \multicolumn{2}{c}{Baseline} & 154.6& 100\%& \color{blue}{102.8} & 73 & \color{red}{\textbf{82}} & 36.4 & \color{red}{\textbf{73.6}} \\
        ~ & 2 & 90\% & 29.7  & 19\% & 87.9 & 62 & 75 & 36.3 & 65.3  \\ 
        ~ & 2 & 75\% & 50.2  & 32\% & 100.5 & 72.5 & 80.9 & \color{red}{\textbf{38.1}} & 73.0  \\ 
        ~ & 2 & 50\% & 84.6  & 55\% & \color{red}{\textbf{103.1}} & \color{blue}{73.4} & 81 & \color{blue}{36.7} & \color{red}{\textbf{73.6}}  \\ 
        ~ & 3 & 90\% & 33.0  & 21\% & 90.2 & 63.6 & 75.2 & 34.9 & 66.0  \\ 
        ~ & 3 & 75\% & 52.9  & 34\% & 100.9 & 72.1 & 79.5 & 36.4 & 72.2  \\ 
        ~ & 3 & 50\% & 86.4  & 56\% & 102.7 & \color{blue}{73.4} & \color{blue}{81.3} & 36.4 & \color{blue}{73.5}  \\ 
        ~ & 5 & 90\% & 39.6  & 26\% & 93.5 & 67.4 & 75.8 & 35.4 & 68.0  \\ 
        ~ & 5 & 75\% & 58.4  & 38\% & 101.4 & 72.5 & 80 & 36.2 & 72.5  \\ 
        ~ & 5 & 50\% & 90.1  & 58\% & 102.5 & \color{red}{\textbf{73.5}} & 81.2 & 36.6 & \color{blue}{73.5} \\ 

\midrule
\multirow{4}{*}{QwenVL-Chat-7B} 
        ~ & \multicolumn{2}{c}{Baseline} & 71.9 &100\% & \color{red}{\textbf{94.9}} & \color{red}{\textbf{72.5}} & \color{red}{\textbf{75.6}}  & \color{red}{\textbf{35.8}}&\color{red}{\textbf{69.7}}\\
        ~ &2 & 90\% & 15.8  & 22\% & 81.9 & 61.5 & 68.5 & 35.3 & 61.7  \\ 
        ~ &2 & 75\% & 24.4  & 34\% & 90.5 & 67.0 & 75.1 & 35.3 & 67.0  \\ 
        ~ &2 & 50\% & 39.5  & 55\% & \color{blue}{94.4} & \color{blue}{71.4} & \color{blue}{75.3} & \color{blue}{35.6} & \color{blue}{69.2} \\



\bottomrule
\end{tabular}
}
\label{tab:image_understanding}
\end{table*}

\begin{table*}[!h] 
    \centering
    \caption{Experiments with more models and benchmarks.}
    \resizebox{0.9\textwidth}{!}{
    \begin{tabular}{lccccc}
    \toprule
          Methods             &AI2Diagram $\uparrow$ & SciQA-IMG $\uparrow$  &SeedBench $\uparrow$ &MMVet$\uparrow$ & MME $\uparrow$\\
    \midrule    
         LLaVA-1.5-13B          &  \textbf{59.45}         & 72.99                                                                 &  \textbf{68.23}&30.55&    1827.75                            \\
         + FastV (K=2,R=50\%)                   & 58.96          & \textbf{73.23}                                                    &  68.03    & \textbf{31.25}     &\textbf{1849.68}                        \\
        \midrule
         InstructBLIP-Vicuna-13B & \textbf{45.46} & 61.15 & \textbf{52.11} & \textbf{24.19} & \textbf{1143.5}
         \\
          + FastV (K=2,R=50\%) & 43.12 & 61.23 & 50.41 & 22.15 & 1129.8
         \\
          + FastV (K=5,R=50\%) & 44.39 & \textbf{62.33} & 51.69 & 23.51 & 1140.5
         \\

    \bottomrule 
    \end{tabular}
    \label{tab:quantitative_results}
    }
\end{table*}

\begin{table*}[!h] 
    \centering    
    \caption{Fine-grained results on MME benchmark.}
    \resizebox{0.9\textwidth}{!}{
    \begin{tabular}{lccccccccccccccccc}
    \toprule
         Methods          & Exist. & Count   & Position & Color & OCR & Poster & Celeb. & Scene & Landmark & Art. & Comm. & Num. & Text. & Code. & Total \\
    \midrule
         LLaVA-1.5-13B    &\textbf{185.00}     &\textbf{155.00}    &\textbf{133.33}   &170.00 &125.00                     &\textbf{160.72}  & 152.54           &161.25        &\textbf{170.50 }  &\textbf{118.50}   &\textbf{128.41 } & \textbf{42.50 }&77.50 & 47.50  & 1827.75 \\
          + FastV (K=2,R=50\%)             &\textbf{185.00}     &\textbf{155.00 }   &\textbf{133.33 }  &\textbf{175.00} &\textbf{132.50 }  &159.77         &\textbf{153.15}          &\textbf{161.75 }  &168.25          &117.00            &126.43             &\textbf{42.50}  &\textbf{82.50 }  &\textbf{57.50 } & \textbf{1849.68}\\
         
    \bottomrule
    \end{tabular}
    \label{tab:mme}
    }
\end{table*}
\subsection{Evaluation Tasks}

We conduct a wide range of evaluation including image captioning, VQA, multimodal reasoning, video QA and fine-grained benchmarks like MME~\cite{fu2023mme} to examine the influence of FastV on the performance of LVLMs. We use greedy search for all experiments and provide details for each task in section A in the supplement material.

\subsection{Model Settings}

We test FastV with various open source models. For image understanding tasks, we conduct experiments on LLaVA1.5-7B, 13B~\cite{liu2023llava15}, and Qwen-VL~\cite{Qwen-VL}. When it comes to video understanding tasks, our baseline model is VideoLLaVA~\cite{lin2023video}. We adopt the settings  as reported in their paper for the baseline models.

\subsection{Main Results}

\paragraph{Image Understanding.} The performance on tasks under different FastV settings are shown in Table~\ref{tab:image_understanding} (Nocaps, Flickr30k, A-OKVQA, MMMU) and Table~\ref{tab:pca} (PCA-Bench, OCR-VQA). The result of latency test is shown in Table~\ref{tab:latency}. 

In Table~\ref{tab:image_understanding}, we present the performance trend with FLOPs ratio ranging from 19\% to 100\% by FastV, for different type and size of models. We also plot the relation between FLOPs Reduction ratio (1-FLOPs Ratio) and average performance in Figure~\ref{fig:pareto_front}. The results indicate that FastV (K=2, R=50\%) could achieve about 45\% FLOPs reduction for different LVLMs without sacrificing the performance. The FLOPs-Performance trade-off is is also highly adjustable by lowering $K$ and increasing $R$ if we want to pursue an ultimate speed up. As shown in the latency test (Table~\ref{tab:latency}), an 13B model with FastV could inference as fast as a 7B model with superior performance for A-OKVQA.

In PCA-Bench and OCR-VQA, (Table~\ref{tab:pca}), which runs finegrained analysis on  perception, cognition, action and OCR abilities, we find that FastV (K=2, R=50\%) could maintain the sub-scores while significantly decreasing the FLOPs.

\begin{table}[!t]
    \caption{Real inference budget comparison between FastV and vanilla decoding. To get rid of the influence of output sequence length on decoding time, we report the result on A-OKVQA dataset where the model only needs to output an option. With FastV, an 13B model could inference as fast as a 7B model while maintaining its superior performance. The latency experiments are conducted on single A40 GPU. }
    \centering
    \resizebox{0.9\textwidth}{!}{
    \begin{tabular}{lccccc}
    \toprule
        Model & Total-Time & GPU-Memory &  Score & Latency/Example \\ 
        \midrule
        LLaVA-1.5-7B & 6:34 & 19G & 76.7 & 0.344s \\ 
        \quad w/ FastV (K=0, R=50\%) & 4:23 & 16G & 75.3 & 0.230s \\ 
        LLaVA-1.5-13B & 10:17 & 38G & 82.0 & 0.539s \\ 
        \quad w/ FastV  (K=0, R=50\%) & 6:30 & 30G & 80.5 & 0.341s \\ 
        \bottomrule
    \end{tabular}}

    \label{tab:latency}
\end{table}

\begin{table}[!t]
    \caption{Finegrained Results on PCA-Bench and OCR-VQA. P, C, and A each denotes Perception, Cognition and Action score. G-PCA denotes Genuine PCA score where the model must make correct perception, cognition and action for one test example to gain 1 score.   The scores are averaged among all three domains including Auto-Driving, Domestic Robot and Open-World Game.}
    \centering
    \resizebox{0.9\columnwidth}{!}{
    \begin{tabular}{l|c|cccc|cccc|c}
    \toprule
        \multirow{2}{*}{Model} &\multirow{2}{*}{FLOPs}& \multicolumn{4}{c|}{PCA-Bench Open Test} & \multicolumn{4}{c|}{PCA-Bench Closed Test} & OCRVQA  \\ 
        ~&~&P & C & A &  G-PCA & P & C & A &  G-PCA & Rouge-L\\
        \midrule
        LLaVA-1.5-7B &99.3B & 0.493 & 0.353 & 0.433 & 0.263 & 0.513 & 0.387 & 0.450 & 0.277 & 0.51 \\ 
        LLaVA-1.5-13B &154.6B &0.530 & \textbf{0.460} & 0.503 & 0.333 & 0.563 & \textbf{0.550} & 0.573 & 0.353 & \textbf{0.55} \\ 
        \quad w/ FastV  (K=0, R=50\%)&78.9B & 0.490 & 0.395 & 0.443 & 0.292 & 0.519 &0.450 & 0.512& 0.283 &0.49\\ 
        \quad w/ FastV (K=2, R=50\%) &84.6B& \textbf{0.533}  & 0.423 & \textbf{0.513} & \textbf{0.340} & \textbf{0.581}&0.545&\textbf{0.580} &\textbf{0.368} & \textbf{0.55} \\
        \quad w/ FastV (K=2, R=75\%) &\textbf{50.2B} &0.513 & 0.417 & 0.483 & 0.320 & 0.523 & 0.510 & 0.533 & 0.323  &0.54 \\ 

        \bottomrule
    \end{tabular}}

    \label{tab:pca}
\end{table}

\paragraph{Video Understanding.} The results of FastV on different video question answering tasks in shown in table~\ref{tab:video_result} (TGIF, MSVD, MSRVTT). To our surprise, we find FastV could generally improves the Video-QA tasks performance while saving 40\%+ computations especially for the TGIF task. We think the main reason is that the redundancy information problem is more severe for video understanding as multiple images from the video are transformed to tokens when sending to the LLM. For example, an image costs 576 tokens in LLaVA1.5 model, while a video costs 2048 tokens in Video-LLaVA. As shown in the case from Figure~\ref{fig:fastv}, setting suitable FastV parameters could lead to much FLOPs reduction for Video-LLaVA while the outputs are nearly identical. 

\paragraph{Fine-grained Benchmarks and More Models} We conduct additional experiments with InstructBLIP and also with more fine-grained LVLM benchmarks such as SciQA-IMG\cite{lu2022scienceqa}, SeedBench~\cite{li2023seedbenchbenchmarkingmultimodalllms}, MMVet~\cite{yu2023mmvetevaluatinglargemultimodal}, and MME~\cite{fu2023mme}, together with benchmarks requiring more visual processing such as AI2Diagram. The results and fine-grained scores of MME are shown in Table~\ref{tab:quantitative_results} and Table~\ref{tab:mme}. FastV works well on different LVLM benchmarks with competitive performance. We find that InstructBLIP shows slightly more performance degradation than LLaVA with same FastV config. The gap soon closes when we just set K to 5. We think it's because Q-Former initially reduces image tokens, resulting in direct information loss. Consequently, it requires adjusting the FastV parameters to avoid too much information loss.

\begin{table}[!t]
    \caption{GPT-Evaluation Results on Video Question Answering Tasks.}
    \centering
    \resizebox{0.9\columnwidth}{!}{
    \begin{tabular}{l|cc|cc|cc|cc}
    \toprule
        \multirow{2}{*}{Model} & \multicolumn{2}{c|}{TGIF} & \multicolumn{2}{c|}{MSVD} & \multicolumn{2}{c|}{MSRVTT} & \multicolumn{2}{c}{Avg} \\ 
        ~&Acc & Score & Acc &  Score & Acc & Score & Acc & Score\\
        \midrule
        Video-LLaVA (Flops=100\%) & 0.18 & 2.5 & 0.70 &3.9 & 0.56 & \textbf{3.5} &0.48 & 3.3   \\ 
        \quad w/ FastV (K=2, R=50\%, Flops=52.3\%) & \textbf{0.21} & \textbf{2.6} & \textbf{0.71} & 3.9 & 0.55 & \textbf{3.5} &\textbf{0.49} & 3.3  \\
        \quad w/ FastV (K=5, R=50\%, Flops=57.1\%) & 0.20 & \textbf{2.6} & \textbf{0.71} & \textbf{4.0}&\textbf{0.57} &\textbf{3.5} & \textbf{0.49} & \textbf{3.4} \\

        \bottomrule
    \end{tabular}}

    \label{tab:video_result}
\end{table}

\begin{table}[!t]
    \caption{Ablation studies results. Scores labelled as ``Failed'' denotes the model could not follow instructions to generates valid results for evaluation.}
    \centering
    \resizebox{0.9\textwidth}{!}{
    \begin{tabular}{l|c|c|c|c}
    \toprule
        Model & Nocaps & Flickr30k & A-OKVQA & MMMU \\ \midrule
        LLaVA1.5-7B (Retrained) & 100.3 & 70.2 & 78.5 & 34.5 \\\midrule 
          (a) w/ Train with 50\% image tokens & 98.5 & 68.5 & 76.8 & 33.5 \\ 
          (b) w/ FastV (K=2, R=50\%)& \textbf{100.1} & \textbf{70} & \textbf{78.4} & \textbf{34.6} \\
          (c) w/ FastV (K=2, R=50\%, Random)& 99.5 & 68.3 & 78.2 & 34.2 \\
          (d) w/ FastV (system prompt) & 89.2 & 64.3 & 69.2 & 33.8 \\
          (e) w/ FastV (prune first half system prompt) & 17.5 & 27.8 & Failed & Failed \\
          (f) w/ FastV (instruction) & 77.3 & 50.1 & 56.5  & 29.5 \\
          (g) w/ StreamingLLM~\cite{xiao2023streamingllm} & 13.2 & 21.4 & Failed & Failed\\
    \bottomrule
    \end{tabular}}
    \label{tab:ablation}
\end{table}


\subsection{Ablation Studies}
\label{sec:ablation}

\begin{figure}[h]
\centering
\includegraphics[width=0.4\textwidth,keepaspectratio]{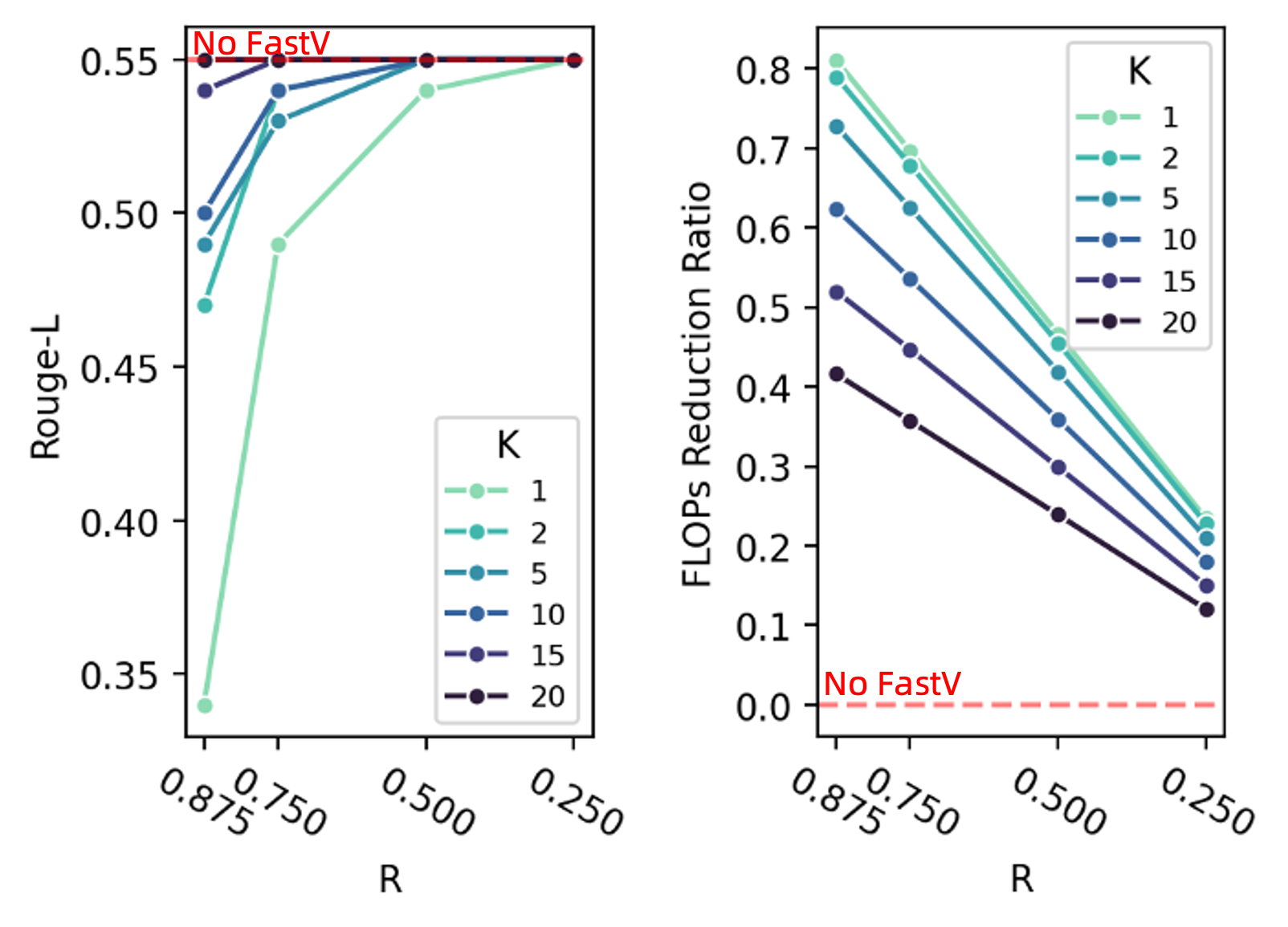}
\caption{Ablation study on filtering layer $K$ and filtering ratio $R$ in FastV. Experiments are conducted with LLaVA1.5-13B on OCR-VQA task. When K is small, lowering R would improve the performance with a smaller FLOPs reduction ratio. In contrast, when K is large, changing R has minimal impact on the overall performance.}
\label{fig:kr-ablation}
\end{figure}

\paragraph{Balance between Cost and Performance.}  We conduct an ablation experiment on how the parameters (K and R) influence the acceleration and downstream task's performance. We select OCR-VQA as the task, which necessitates a through understanding of the image. The result is shown in Figure~\ref{fig:kr-ablation}. \textbf{When K is small, lowering R would improve the performance with a smaller FLOPs reduction ratio. In contrast, when K is large, adjusting R has minimal impact on the overall performance.} This observation further proves that in deep layers, there is high redundancy in image tokens.

\paragraph{Training with Less Tokens.}
FastV reduces computational requirements (FLOPs) by pruning tokens during the inference stage. An alternative approach for token reduction involves training the LVLM at a lower resolution. To facilitate a fair comparison, we retrained two LLaVA1.5-7B models, adhering to the original pretraining and supervised finetuning protocols. The sole modification in the second model's training process was the incorporation of an average pooling layer (with a stride of 2) following the Clip encoder, leading to a 50\% reduction in image tokens during training. A comparison between lines (a) and (b) in Table~\ref{tab:ablation} reveals that reducing the input resolution directly during training results in diminished performance. Conversely, FastV manages to decrease the number of image tokens without compromising performance, showcasing its efficiency in balancing computational savings with model efficacy.

\paragraph{Pruning Token Strategy.} FastV strategically reduces the number of image tokens during the inference phase of LVLMs, motivated by our observation that image tokens exhibit the lowest attention efficiency relative to other types of input tokens. In experiments detailed in lines (d) and (f) of the study, we specifically pruned tokens that were not related to images, such as system prompts and instruction tokens. This selective pruning resulted in significant performance declines, even when only a minimal number of non-image tokens were removed. We also compare randomly drop visual tokens instead of dropping by attention rank, as shown in line (c). It resulted in declined results compared with origin FastV (b). These findings underscore the distinct roles that visual and textual tokens play within LVLMs. It highlights FastV's effectiveness in precisely targeting image tokens for reduction, thereby optimizing performance without compromising the model's overall functionality. 

In our previous observation about attention efficiency, we find out that the system prompt takes up of most attention even if they carry the least semantic information in the context. We conduct another experiment by directly prune the first half tokens of the system prompt. Comparing line (d) and (e), we can find that the head tokens in the system prompt have dominant effect on the model performance. Our findings also align with StreamingLLM~\cite{xiao2023streamingllm} where they find that the first 4 tokens in LLM play the most important role during inference. However, direcly applying the same sparse attention pattern as StreamingLLM would lead to a substantial degradation in LVLM's performance as shown in line (g) of Table~\ref{tab:ablation}. This suggests a fundamental difference in how image tokens, as opposed to text tokens, contribute to the information processing within LLMs.




\section{Conclusion}
In this paper, we propose FastV, a plug-and-play inference cost optimization method for Large Vision-Language Models. Our insight for FastV arises from our observation that the attention computation over visual tokens is of extreme inefficiency in the deep layers of popular LVLMs though they take up a large portion of input tokens. FastV prunes out the unnecessary visual tokens according to the attention score ranking, which results in significant inference cost reduction without sacrificing performance.

\newpage

\bibliography{colm2024_conference}
\bibliographystyle{colm2024_conference}

\appendix
\section{Appendix}
\begin{figure}[!t]
\centering
\includegraphics[width=0.7\textwidth,keepaspectratio]{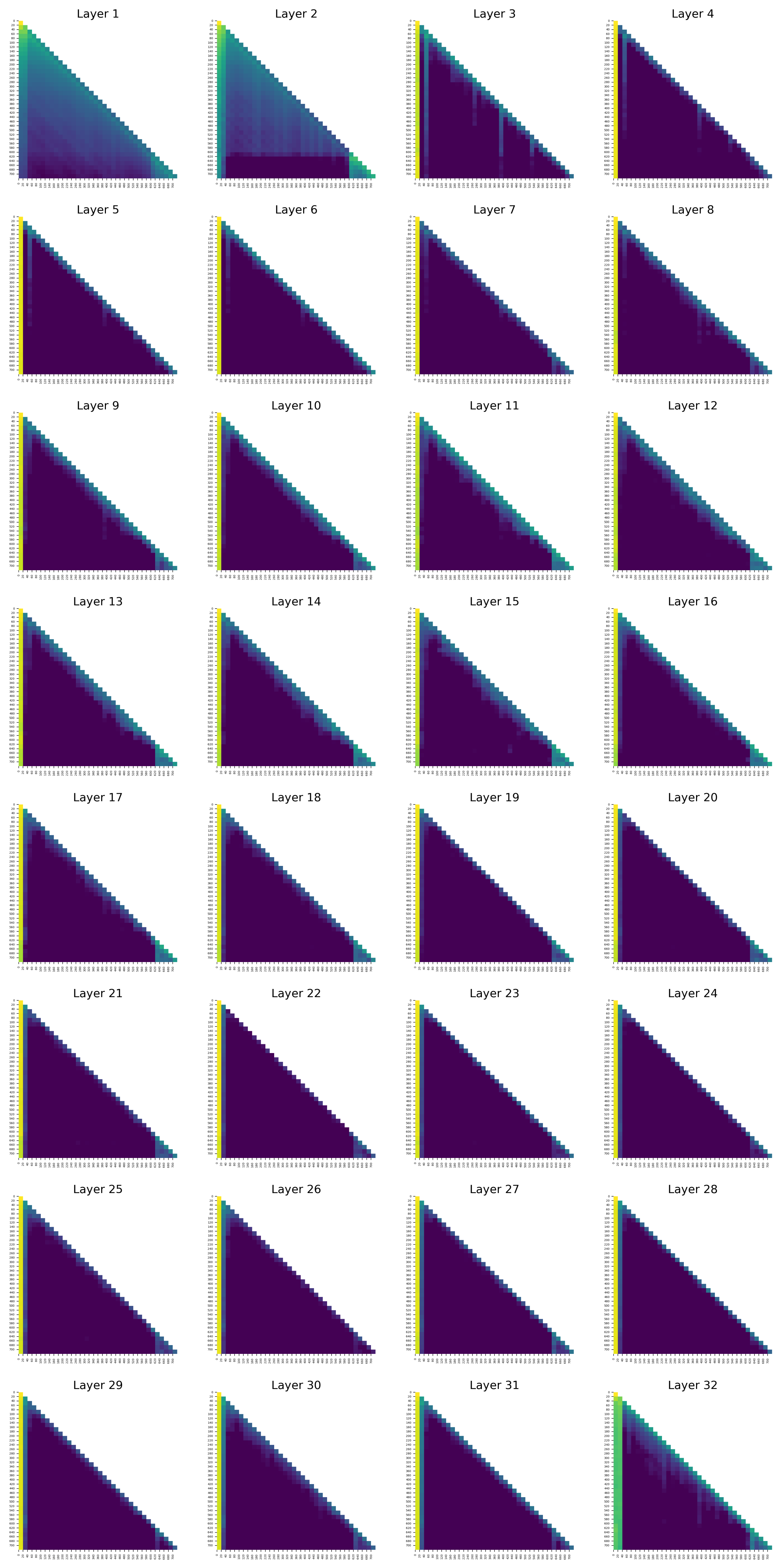}
\caption{Full Attention Maps of Each Layer of LLaVA.}
\label{fig:full_attention_map}
\end{figure}

\section{Evaluation Tasks Description}
\label{app:task description}

\paragraph{Image Captioning.} Image captioning requires the model to generate a description for a given image. We choose Nocaps~\cite{agrawal2019nocaps} and Flickr30k~\cite{plummer2015flickr30k} as benchmarks and report CIDEr score~\cite{vedantam2015cider} as metric.  For image captioning tasks Nocaps and Flickr30k, we adopt prompt as ``Describe the image in one sentence.''

\paragraph{Visual Question Answering (VQA).} VQA  requires the model to generate an answer for a given image-question pair. We select the development set of A-OKVQA~\cite{schwenk2022aokvqa} and the test set of OCR-VQA ~\cite{mishra2019ocr_vqa} as the benchmark and the report the multiple choice (MC) score of AOKVQA and Rouge-L score of OCR-VQA.  For AOKVQA, we adopt the the multiple choice version of evaluation and use prompt as: ``Analyse the image and choose the best answer for the following question: \{question\} Options: \{options\}. Output the letter of the correct answer.'' For OCRVQA, we use the default question as prompt for each example as provided in the official dataset.

\paragraph{Multimodal Reasoning.}
Compared with VQA, multimodal reasoning requires more advanced perception, knowledge and reasoning skills of the model, which are more suitable benchmarks to evaluate the integrated abilities of LVLMs. We choose MMMU and PCA-Bench~\cite{chen2024pcabench} as benchmarks. MMMU is a multimodal benchmark featuring multi-discipline tasks
demanding college-level subject knowledge and reasoning skills. PCA-Bench is a complex embodied reasoning benchmark with error localization, which features three different domains including autonomous driving, robot and game. We report the multiple choice accuracy for the development set of MMMU and Perception, Cognition, Action, Genuine PCA scores for both the open and closed test set of PCA-Bench. We use the default prompts for each example as provided in the official dataset \href{https://huggingface.co/datasets/MMMU/MMMU}{MMMU} and \href{https://huggingface.co/datasets/PCA-Bench/PCA-Bench-V1}{PCA-Bench}.

\paragraph{Video Question Answering.} Similar to VQA for single image, Video Question Answering requires the model to generate answer given a video-question pair. Current LVLMs usually deal with video question answering tasks by sampling multiple frames as input, resulting in longer image token sequences. We choose TGIF-QA~\cite{jang2017tgifqa}, MSVD-QA~\cite{xu2017msvdqa} and MSRVTT-QA~\cite{xu2017msrvtt_qa} as benchmarks following the evaluation pipeline of Video-ChatGPT~\cite{Maaz2023VideoChatGPT} and report the accuracy and chatgpt-score as metrics. We use the first 1K examples in each benchmark in our experiments due to the limited commercial API usage in evaluation. For all video QA tasks, we use the default question as the prompt as provided in \href{https://github.com/PKU-YuanGroup/Video-LLaVA/blob/main/TRAIN_AND_VALIDATE.md}{Video-LLaVA}, and use the same tool from \href{https://github.com/mbzuai-oryx/Video-ChatGPT/blob/main/video_chatgpt/eval/run_inference_benchmark_general.py}{Video-ChatGPT} to conduct GPT evaluation.

\paragraph{Fine-grained Benchmarks}
For the evaluation of the influence of FastV on LVLM performance, we incorporate four distinct Fine-grained benchmarks: MME~\cite{fu2023mme}, Seed-Bench~\cite{li2023seedbenchbenchmarkingmultimodalllms}, SciQA-IMG~\cite{lu2022scienceqa}, and MMVet~\cite{yu2023mmvetevaluatinglargemultimodal}.
MME offers a comprehensive evaluation of models' perception and cognition abilities across a diverse set of tasks, focusing on intuitive and quantifiable analysis without extensive prompt engineering. SEED-Bench, on the other hand, evaluates generative comprehension across multiple dimensions, ensuring question relevance and quality through a mix of automated filtering and manual verification.
While  MME and SEED-Bench cover general abilities of LVLMs, SciQA-IMG and MMVet focus on the advanced aspects of multi-modal understanding. SciQA-IMG is a large-scale multimodal science question dataset annotated with detailed lectures and explanations. MMVet evaluates LVLMs on complex multimodal tasks, emphasizing multi-modal understanding and free-form answering capabilities, thus offering a comprehensive view of model performance.

\end{document}